\documentclass[conference]{IEEEtran}
\IEEEoverridecommandlockouts

\usepackage{cite}
\usepackage{amsmath,amssymb,amsfonts}
\usepackage{graphicx}
\usepackage{booktabs}
\usepackage{multirow}
\usepackage{textcomp}
\usepackage[T1]{fontenc}
\usepackage[utf8]{inputenc}
\usepackage{xurl}  
\usepackage{xcolor}
\usepackage{tikz}
\usetikzlibrary{arrows.meta,positioning,shapes.geometric,fit,backgrounds}
\usepackage{microtype}
\usepackage[hidelinks,breaklinks]{hyperref}
\hypersetup{pdftitle={Explainable Detection of Anomalous Unit Prices in Public Procurement: Robust Screening and Learned Ranking on Annotated Verdicts},
  pdfauthor={Dmytro Babarytskyi, Tetiana Bilova, Oleksandr Babarytskyi}}

\graphicspath{{figs/}}
\newcommand{\doi}[1]{\href{https://doi.org/#1}{doi:~\detokenize{#1}}}

\begin{document}

\title{Explainable Detection of Anomalous Unit Prices in Public Procurement:
Robust Screening and Learned Ranking on Annotated Verdicts}

\author{\IEEEauthorblockN{Dmytro Babarytskyi}
\IEEEauthorblockA{\textit{Kharkiv National University}\\
\textit{of Radio Electronics}\\
Kharkiv, Ukraine\\
dmytro.babarytskyi@nure.ua\\
ORCID: 0009-0009-3614-9601}
\and
\IEEEauthorblockN{Tetiana Bilova}
\IEEEauthorblockA{\textit{Kharkiv National University}\\
\textit{of Radio Electronics}\\
Kharkiv, Ukraine\\
tetiana.bilova@nure.ua\\
ORCID: 0000-0002-1085-7361}
\and
\IEEEauthorblockN{Oleksandr Babarytskyi}
\IEEEauthorblockA{\textit{Military Law Institute}\\
\textit{Yaroslav Mudryi National}\\
\textit{Law University}\\
Kharkiv, Ukraine\\
aleksandr.babaritskiy@ukr.net\\
ORCID: 0000-0002-1690-1837}
}

\maketitle

\begin{abstract}
Public e-procurement systems publish millions of purchased items with their unit prices, yet price
screening in audit practice still relies on manual comparison or on procedure-level ``red flags'' that
ignore what was actually bought and at what price. We present a two-stage method for detecting anomalous
unit prices at the level of individual procured items and explaining each decision to an auditor. The
first stage makes heterogeneous item records comparable --- by converting per-package prices into prices
per kilogram or litre, grouping items hierarchically by procurement classification code and unit, and
splitting large groups by the semantics of their free-text descriptions --- and then screens each group
with robust statistical and density-based rules. The second stage is a monotone gradient-boosting ranking
model that re-scores the screening signals by how likely an annotator is to confirm the item; its
decisions are explained per item with Shapley values in auditor-readable terms. We evaluate the method on
16{,}103 food items from the Ukrainian Prozorro system with a stratified, design-weighted sample of 400
items. The labels were produced by a large language model under a written protocol and checked by a
non-expert human on 80 items ($\kappa=0.64$); they are not verdicts of professional auditors. A first
annotation round that displayed the earlier system's decision confirmed its flags more often than a blind
round, a shift consistent with anchoring ($\kappa=0.82$ between the rounds). On blind labels, the ranking
model raises the design-weighted average precision (AP) from 0.49 for the screening of the first system
version and 0.51 for the best statistical rule to 0.81 (95\% bootstrap interval of the gain over the rule:
0.10--0.44); with the human corrections the gain over the rule shrinks to 0.09 and is no longer significant
($-0.09$--0.30). On injected errors,
flagging gross extremes independently of the ranking model is needed to keep the recall of unit-scale errors.
Explanations computed with 24 Monte Carlo permutations are within about one point of a 2{,}000-permutation
reference on a 0--100 risk scale. The method is implemented in an open-source information system.
\end{abstract}


\section{Introduction}

Public procurement accounts for a large share of government spending, and its digitalisation has made
transaction-level data openly available. In the Ukrainian e-procurement system Prozorro alone,
207 thousand procurements worth UAH~463 billion (UAH: Ukrainian hryvnia) were announced in the fourth quarter of 2025~\cite{prozorro2026q4}; each
procurement contains one or more items with a description, a classification code under the Common
Procurement Vocabulary (CPV, national version DK~021), a unit of measure, a quantity and --- in contracts or
in the expected value --- a price. This openness creates an opportunity for continuous, data-driven audit:
an auditor could, in principle, compare every purchased item with every comparable purchase.

In practice, item-level price screening is hard. Records that should be comparable are not: the same good
is bought per kilogram, per piece or per pack of unspecified weight, descriptions are free text, and a
single CPV code can cover goods whose prices differ by an order of magnitude. Data errors (a lot total
entered as a unit price, grams confused with kilograms) are common --- in our annotated sample they make up
31 of the 54 confirmed items --- and are relevant for an auditor, because they can hide or simulate
over-pricing and make published spending figures unreliable. Existing
automated risk indicators --- including those used by the Ukrainian State Audit Service~\cite{minfin2024indicators}
and the corruption risk indices of the procurement literature~\cite{fazekas2016objective,gti2024indicators}
--- operate at the level of procedures (single bidding, short tender periods, repeated winners) and largely
leave the price of what was bought unexamined.

Machine learning for procurement oversight has mostly been supervised on procedure-level outcomes, such as
sanctions, collusion cases or malfeasance reports~\cite{gallego2021preventing,huber2019screens,tas2024collusion},
or unsupervised over contract-level features~\cite{niessen2020ocds,torresberru2021mining}. Two obstacles
recur: labels are scarce and costly, and a flag without an explanation is hard for an auditor to act on.

This paper addresses item-level price screening as a decision-support problem with an auditor in the
loop. Our contributions are:
\begin{enumerate}
  \item \textbf{A comparability layer for item-level procurement prices}: pack-size parsing that converts
  per-piece prices into prices per kilogram or litre, hierarchical grouping by unit and CPV code with
  bottom-up aggregation of small groups, and semantic splitting of large groups by their descriptions.
  \item \textbf{A two-stage detector}: unsupervised screening within comparable groups (Tukey extremes,
  robust $z$-scores, DBSCAN in a joint price--quantity--season space) followed by a monotone ranking model
  trained on annotation labels, plus a separate data-error signal. The model's features are
  product-agnostic, which makes it applicable to other categories; on injected errors in a held-out
  category it remains competitive in F1 but ranks errors worse than screening.
  \item \textbf{Item-level explanations} with Monte Carlo Shapley values over auditor-readable features,
  with an evaluation of their accuracy, stability and fidelity.
  \item \textbf{An evaluation protocol for label-scarce audit settings}: stratified annotation with
  Horvitz--Thompson weighting, a blind re-annotation that shows a shift, consistent with anchoring, between a
  sighted and a blind round, a check of language-model labels by a human, and a synthetic benchmark of error types
  typical of procurement data.
\end{enumerate}
The method is implemented in an information system that synchronises Prozorro data, runs the detector on
auditor-defined selections, and records verdicts with four-eyes approval; reviewed verdicts can be used to
retrain the ranking model.

\section{Related work}

\paragraph{Risk indicators in public procurement} Corruption risk research has developed composite
indices from procedure-level red flags~\cite{fazekas2016objective,gti2024indicators}. Their predictive
value is contested: red flags that are known in advance can be gamed, and the most scrutinised indicators
may be the least informative~\cite{decarolis2022redflags}. Ukrainian practice combines official automatic
indicators~\cite{minfin2024indicators} with civil-society tools such as DOZORRO, whose machine-learning
component scores procedures for likely violations~\cite{tiu2018dozorro}. None of these indicators targets
the unit price of the purchased item. Economic studies do compare unit prices, but for a different purpose:
they restrict attention to homogeneous goods and estimate how much public bodies overpay on average, as a
measure of waste~\cite{bandiera2009waste} or of the effectiveness of procurement officials and
organisations~\cite{best2023effectiveness}. They do not flag individual items for review and work with goods
whose units are already comparable.

\paragraph{Machine learning for procurement oversight} Supervised models predict malfeasance from
procedure and municipality characteristics~\cite{gallego2021preventing}, detect bid rigging from
statistical screens of bid distributions~\cite{huber2019screens,tas2024collusion}, or predict corruption
indices~\cite{rabuzin2019prediction}. Unsupervised approaches apply clustering and outlier detection to
contract features~\cite{niessen2020ocds,torresberru2021mining,herreros2025purchase}. These works typically
treat a contract or lot as the unit of analysis; item descriptions and units of measure, which determine
whether two prices are comparable at all, receive little attention.

\paragraph{Unsupervised anomaly detection} Robust univariate rules (Tukey fences~\cite{tukey1977eda},
the modified $z$-score~\cite{iglewicz1993outliers}) remain strong baselines for one-dimensional price data.
Density- and isolation-based detectors --- DBSCAN~\cite{ester1996dbscan}, the local outlier factor
(LOF)~\cite{breunig2000lof}, Isolation Forest~\cite{liu2008iforest} and empirical-cumulative-distribution
outlier detection (ECOD)~\cite{li2023ecod} --- extend screening to several attributes;
see~\cite{chandola2009survey} for a survey. Their output is a score with no notion of what a domain
expert would consider worth investigating.

\paragraph{Explanations and the human in the loop} Shapley values~\cite{shapley1953value} provide
additive feature attributions with axiomatic guarantees; they are estimated by permutation sampling
for arbitrary models~\cite{strumbelj2014explaining} or computed efficiently for trees~\cite{lundberg2017shap,lundberg2020trees}.
Large language models have recently been used as annotators in text-classification tasks~\cite{gilardi2023chatgpt};
we use one under a written protocol and verify it against a human subsample. Anchoring on a provided
reference~\cite{tversky1974judgment} is a known threat when annotators see model output, which motivates
our blind re-annotation.

\section{Data and problem setting}\label{sec:data}

Prozorro publishes procurement records through an open API~\cite{prozorroapi} following the Open
Contracting Data Standard~\cite{ocds}. We synchronise the change feed into a document store and normalise
it into a relational schema of tenders, lots, items, contracts, procuring entities and suppliers. For every
item we derive a unit price from, in order of preference, an item-level unit value, the contract, or the
expected value, and we convert the unit of measure to a base unit (kilogram, litre, piece, \ldots) using
UN/CEFACT codes.

\begin{table*}[t]
\centering
\caption{Data synchronised from Prozorro (change feed of 10--24 September 2026). The reference selection
filters publication dates between 1 August and 30 September 2026; because only the change feed of
10--24 September was synchronised, 93\% of its items were published on 10--24 September.}
\label{tab:data}
\small
\begin{tabular}{lrrrr}
\toprule
CPV division & Tenders & Items & Items with price and quantity & \ldots{} in reference period \\
\midrule
15 Food, beverages & 18{,}824 & 33{,}527 & 16{,}117 & 16{,}103 \\
03 Agricultural products & 7{,}850 & 21{,}266 & 6{,}511 & 6{,}507 \\
\bottomrule
\end{tabular}
\end{table*}
Table~\ref{tab:data} summarises the data. Unit prices of CPV~15 items come from contracts (66\%),
item-level unit values (25\%) and expected values (9\%). CPV~15 is used for annotation and training of the
ranking model; CPV~03 is held out to test transfer. The effective observation window is short: the
synchronised change feed covers 10--24 September 2026, so items published earlier enter only if their
tender changed in that window (153 of the 16{,}103 CPV~15 items were published in August).

The analysis unit is an item with a positive unit price and quantity. The auditor selects CPV codes, a
publication period and optionally units; the detector returns, for every item, a risk score in $[0,100]$,
a binary decision and an explanation. The reference selection used for annotation is CPV division~15
(food products) with the publication filter 1~August--30~September~2026: 16{,}103 items. The first system
version (v1, see below) analysed 15{,}450 of them in 340 groups; the current version (v2) analyses 15{,}034
in 320 groups.

\section{Method}\label{sec:method}

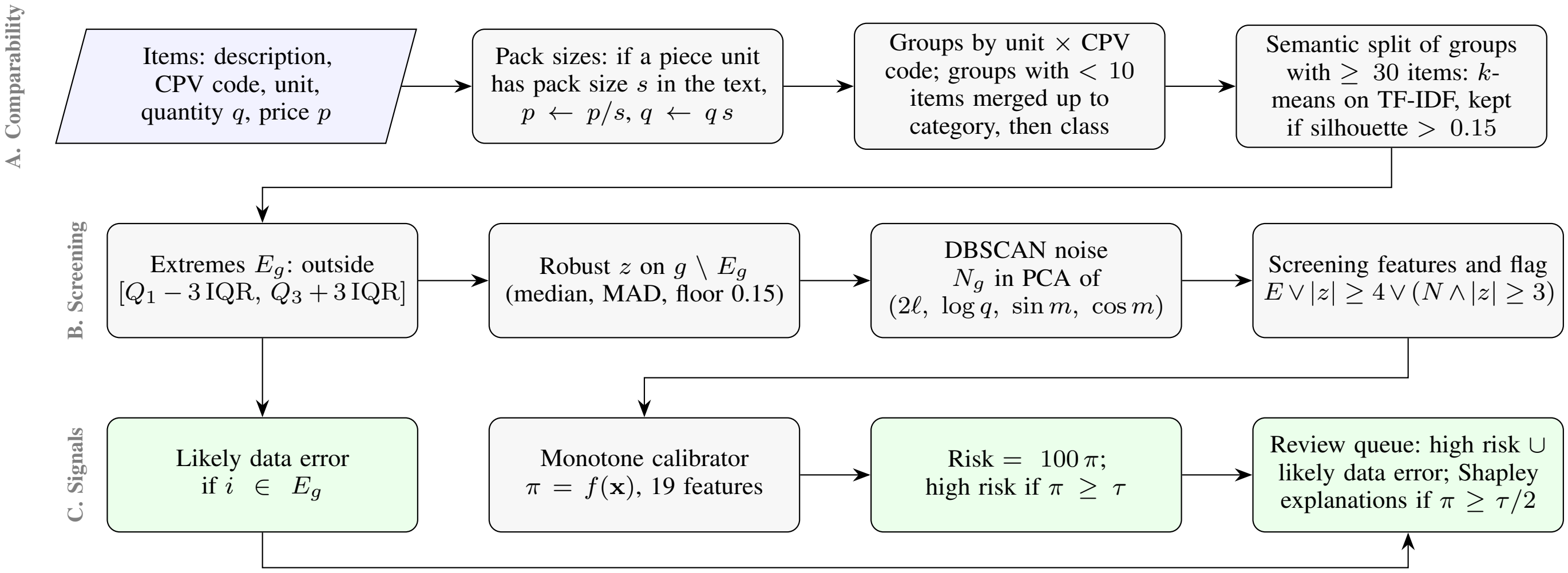
\begin{figure*}[t]
\centering
\resizebox{\textwidth}{!}{%
\begin{tikzpicture}[
  node distance=9mm and 8mm, every node/.style={font=\footnotesize},
  proc/.style={draw, rounded corners, align=center, minimum height=13mm, text width=33mm, fill=gray!6},
  io/.style={draw, trapezium, trapezium left angle=75, trapezium right angle=105, align=center,
             minimum height=13mm, text width=27mm, fill=blue!5},
  outbox/.style={draw, rounded corners, align=center, minimum height=13mm, text width=33mm, fill=green!8},
  stage/.style={rotate=90, anchor=south, font=\scriptsize\bfseries, text=gray},
  arr/.style={-{Stealth[length=2mm]}}]
\node[io] (a1) {Items: description, CPV code, unit, quantity $q$, price $p$};
\node[proc, right=of a1] (a2) {Pack sizes: if a piece unit has pack size $s$ in the text,\\$p\gets p/s$, $q\gets q\,s$};
\node[proc, right=of a2] (a3) {Groups by unit $\times$ CPV code; groups with $<10$ items merged up to category, then class};
\node[proc, right=of a3] (a4) {Semantic split of groups with $\ge30$ items: $k$-means on TF-IDF, kept if silhouette $>0.15$};
\node[proc, below=of a1.south, anchor=north, xshift=3mm] (b1) {Extremes $E_g$: outside\\$[Q_1-3\,\mathrm{IQR},\,Q_3+3\,\mathrm{IQR}]$};
\node[proc, right=of b1] (b2) {Robust $z$ on $g\setminus E_g$\\(median, MAD, floor 0.15)};
\node[proc, right=of b2] (b3) {DBSCAN noise $N_g$ in PCA of $(2\ell,\ \log q,\ \sin m,\ \cos m)$};
\node[proc, right=of b3] (b4) {Screening features and flag\\$E\lor|z|\ge4\lor(N\land|z|\ge3)$};
\node[outbox, below=of b1] (c1) {Likely data error\\if $i\in E_g$};
\node[proc, right=of c1] (c2) {Monotone calibrator\\$\pi=f(\mathbf{x})$, 19 features};
\node[outbox, right=of c2] (c3) {Risk $=100\,\pi$;\\high risk if $\pi\ge\tau$};
\node[outbox, right=of c3] (c4) {Review queue: high risk $\cup$ likely data error; Shapley explanations if $\pi\ge\tau/2$};
\node[stage] at ([xshift=-4mm]a1.west) {A. Comparability};
\node[stage] at ([xshift=-1mm]b1.west) {B. Screening};
\node[stage] at ([xshift=-1mm]c1.west) {C. Signals};
\draw[arr] (a1) -- (a2); \draw[arr] (a2) -- (a3); \draw[arr] (a3) -- (a4);
\draw[arr] (a4.south) -- ++(0,-4.5mm) -| (b1.north);
\draw[arr] (b1) -- (b2); \draw[arr] (b2) -- (b3); \draw[arr] (b3) -- (b4);
\draw[arr] (b1) -- (c1);
\draw[arr] (b4.south) -- ++(0,-4.5mm) -| (c2.north);
\draw[arr] (c2) -- (c3); \draw[arr] (c3) -- (c4);
\draw[arr] (c1.south) -- ++(0,-4mm) -| (c4.south);
\end{tikzpicture}}
\caption{The two-stage method. A: item records are made comparable. B: each group with at least ten items
is screened with robust rules ($\ell=\log p$, $m$ the month). C: a monotone calibrator turns the screening
features into a risk; extremes are reported in addition as likely data errors, independently of the
calibrator; both signals form the review queue.}
\label{fig:pipeline}
\end{figure*}

Fig.~\ref{fig:pipeline} outlines the method: a comparability layer, screening within comparable groups
(the \emph{screening stage}), a learned ranking model that re-scores the screening signals (for brevity, the
\emph{calibrator}; its outputs are scores used for ranking, not population probabilities), a separate
data-error signal, and per-item explanations. We call \emph{v1} the first deployed version, which consisted
of the screening stage without pack-size conversion and with a coarser semantic split, and \emph{v2} the
current version: the full screening stage followed by the calibrator.

\subsection{Comparability layer}

\paragraph{Pack sizes} Items bought per piece, pack, bottle or jar often state the net content in the
description (``sour cream 20\%, 350~g'', ``oil 5~l'', ``tea 20$\times$2~g''). A rule-based parser extracts the
content and converts the price to UAH per kilogram or litre; the quantity is converted accordingly.

\paragraph{Hierarchical groups} Items are grouped by base unit and full eight-digit CPV code. Groups below
a minimum size ($n<10$) are merged upwards to the five-digit category and then to the four-digit class;
items that remain in groups smaller than the minimum are left unanalysed rather than compared with
dissimilar goods.

\paragraph{Semantic split} A CPV code can still mix goods with very different prices (``dairy products''
covers yoghurt and hard cheese). Groups with at least 30 items are split by description: lemmatised
descriptions (lemmatiser of~\cite{korobov2015pymorphy}) are vectorised with term frequency--inverse
document frequency (TF-IDF)~\cite{salton1988weighting}, reduced with truncated singular value
decomposition (SVD) and
clustered with $k$-means~\cite{macqueen1967kmeans}, choosing $k\le 8$ by the silhouette
coefficient~\cite{rousseeuw1987silhouettes}; the split is kept only if the silhouette exceeds 0.15 and every
subgroup has at least five items. Procurement descriptions are short (median three words), domain-specific
and often contain only the product name, its grade and pack size; lemmatisation followed by TF-IDF separates
such texts well and is fast and transparent. Contextual sentence embeddings~\cite{reimers2019sbert} are a
natural alternative for longer descriptions, at a higher cost and with less interpretable groups, which
matters when the auditor must see why two items were compared.

\paragraph{Why these design choices} Three properties of item-level procurement prices drive the design.
First, prices are heavy-tailed and multiplicative --- a data error easily changes a price by three orders of
magnitude --- so all statistics are computed on log prices with robust estimators (the median and the
median absolute deviation, MAD), and
extremes are removed before the spread is estimated so that a single lot total does not inflate it.
Second, groups are small: in v2 the reference selection yields 320 analysed groups with a median of 24
items, and 76\% of them have fewer than 50 items,
which rules out detectors that need many points per group and motivates bottom-up aggregation by the CPV
hierarchy. Third, the same price can be normal for one purchase and suspicious for another depending on
the quantity, and plausibly on the season, which is why density-based screening uses quantity and the
cyclically encoded month in addition to price. With our two-week window the month is nearly constant, so
the seasonal component is untested here.

\subsection{Unsupervised screening}

Within each group $g$ with log prices $\ell_i=\log p_i$ three rules are applied.

\emph{Extremes.} An item is an extreme if its log price lies outside the outer Tukey fences of the group,
\begin{equation}
E_g=\bigl\{\,i\in g:\ \ell_i\notin[\,Q_1-3\,\mathrm{IQR},\ Q_3+3\,\mathrm{IQR}\,]\bigr\},
\label{eq:extremes}
\end{equation}
where $Q_1$, $Q_3$ are the quartiles of $\ell$ in $g$ and IQR the interquartile range. Extremes are
excluded from the estimation of the group's centre and spread.

\emph{Robust $z$-score.} For every item of the group,
\begin{equation}
z_i=\frac{\ell_i-\tilde\ell_g}{\max\left(1.4826\cdot\mathrm{MAD}_g,\ 0.15\right)},
\end{equation}
where the median $\tilde\ell_g$ and the median absolute deviation $\mathrm{MAD}_g$ are estimated on
$g\setminus E_g$. The floor of 0.15 prevents ordinary market variation of about 15\% from being flagged in
very homogeneous groups. The screening risk score is $100\,(1-e^{-|z_i|/3})$.

\emph{Density outliers.} DBSCAN~\cite{ester1996dbscan} is run in the principal component analysis (PCA) projection~\cite{jolliffe2016pca}
of the standardised vector $(2\ell,\ \log q,\ \sin m,\ \cos m)$ --- price weighted twice, quantity and the
cyclically encoded month --- with $\varepsilon$ chosen at the knee of the sorted $k$-distance
curve~\cite{satopaa2011kneedle}; its noise points form the set $N_g$.

The screening decision combines the three rules,
\begin{equation}
r_i=\bigl[i\in E_g\bigr]\ \lor\ \bigl[|z_i|\ge4\bigr]\ \lor\ \bigl[i\in N_g\ \land\ |z_i|\ge3\bigr].
\end{equation}
A density outlier must thus be a \emph{price} outlier, not merely unusual in quantity or date. The thresholds
(the extreme factor, $|z|\ge4$ and $\ge3$, the MAD floor, the price weight) were chosen by a grid search on
injected anomalies in CPV~15 (so the CPV~15 synthetic results below are in-sample for the screening
stage) and are exposed to analysts as parameters with bounds. With the
knee-based $\varepsilon$, DBSCAN has no free parameter that an analyst would have to tune per group;
its hierarchical variant HDBSCAN~\cite{campello2013hdbscan} removes $\varepsilon$ altogether but requires a
minimum cluster size, which is hard to set for groups ranging from ten to several hundred items.

\subsection{Learned ranking on annotated verdicts}

Screening finds deviations but cannot know which deviations are worth an auditor's time. A
histogram-based gradient-boosting classifier~\cite{friedman2001gbm,ke2017lightgbm} (scikit-learn
implementation~\cite{pedregosa2011sklearn}; depth~3, 300 iterations, balanced class weights) maps 19
screening and context features to a score $\pi\in[0,1]$ that ranks items by how likely the annotator is to
confirm them; the risk shown to the auditor is $100\,\pi$. Because the model is trained without design
weights on the stratified sample, where flagged items are over-represented, $\pi$ is a ranking score rather
than a population probability (Section~\ref{sec:res-labels} reports its reliability). The features are: the signed and absolute robust $z$, the log ratio to the group median, extreme
and density-outlier indicators, group size and spread, TF-IDF distance of the description to the group
centroid, an indicator of generic descriptions (``various food products''), pack conversion and a
``price-per-pack'' mismatch feature (how close the log ratio is to the log pack size), unit type, price
source, and log quantity (Table~\ref{tab:features}). None of the features identifies a product, so the
model can be applied to other categories; how well it does so is examined in Section~\ref{sec:results}. The decision threshold $\tau$ is selected by cross-validation; the
default model uses $\tau=0.675$.

\begin{table}[t]
\caption{Calibrator features and their global importance (mean $|\phi|$ over 120 flagged items, risk
points; Section~\ref{sec:res-explain}). $\uparrow$: monotone non-decreasing constraint.}
\label{tab:features}
\centering
\footnotesize
\setlength{\tabcolsep}{3pt}
\begin{tabular}{@{}p{0.30\columnwidth}p{0.52\columnwidth}r@{}}
\toprule
Feature & Meaning & $|\phi|$ \\
\midrule
\multicolumn{3}{@{}l}{\emph{Deviation (screening stage)}} \\
Deviation $|z|$ $\uparrow$ & absolute robust $z$-score of the log price & 69.3 \\
Direction of deviation $z$ & signed robust $z$ (dearer or cheaper) & 3.9 \\
Price ratio & log ratio of the price to the group median & 7.3 \\
Extreme $\uparrow$ & outside the outer Tukey fences, eq.~\eqref{eq:extremes} & 0.0 \\
Density outlier & DBSCAN noise point & 0.0 \\
\midrule
\multicolumn{3}{@{}l}{\emph{Group context}} \\
Group size & log number of comparable items & 2.6 \\
Price spread & MAD of log prices in the group & 2.6 \\
Description distance & TF-IDF cosine distance of the description to the group centroid & 9.0 \\
Generic description & ``various products'', ``as per invoice'' & 0.4 \\
\midrule
\multicolumn{3}{@{}l}{\emph{Units and packs}} \\
Pack converted & price converted from the pack size & 0.4 \\
Pack size stated & pack size found in the description & 0.3 \\
Per-pack price & closeness of the price ratio to the pack size & 0.0 \\
Piece-type unit & piece, pack, bottle, jar, set & 2.1 \\
Weight unit & base unit kilogram & 0.9 \\
Volume unit & base unit litre & 0.1 \\
\midrule
\multicolumn{3}{@{}l}{\emph{Price source and volume}} \\
Item unit value & price stated per unit in the item & 1.0 \\
Contract price & price taken from the contract & 1.6 \\
Expected value & price taken from the expected value & 0.0 \\
Quantity & log quantity in the base unit & 2.8 \\
\bottomrule
\end{tabular}
\end{table}

With few positive labels, an unconstrained tree ensemble can learn non-monotone relations from
idiosyncrasies of the sample --- in our first blind-trained model, a twelvefold over-priced item with a
converted pack size received a lower score than moderate deviations. We therefore impose monotone
constraints: the score is non-decreasing in the absolute robust $z$-score and in the extreme indicator. The
constraint encodes domain knowledge that a larger deviation is never less suspicious, other things equal.
Because the extreme indicator is nearly determined by $|z|$, its constraint is inactive in practice; four
features (extreme, density outlier, per-pack price, expected-value source) receive no attribution in the
explanations (Table~\ref{tab:features}) and are kept only for compatibility of the feature set. In
particular, DBSCAN influences v2 only through the density-outlier feature, which receives no attribution:
it shapes the v1 screening flag $r_i$ but no output of v2, and we did not ablate it separately.

\paragraph{Implementation details} The classifier uses learning rate 0.05, $L_2$ regularisation 1.0 and at
least ten items per leaf. DBSCAN uses $\textit{min\_samples}=5$ and a PCA keeping 90\% of the variance;
truncated SVD is applied only when the vocabulary of a group exceeds 60 terms; subgroups of five to nine
items produced by the semantic split are left unanalysed. In nested cross-validation the decision for an
item is the majority over the five repetitions.

The ranking model is retrained offline when enough new verdicts have accumulated (in our deployment we plan a
semi-annual to annual cycle), rather than continuously: verdicts given while the auditor sees the model's
output may shift towards it, as the sighted annotation round suggests (Section~\ref{sec:res-labels}), and
periodic retraining allows the training set to be audited first.

\subsection{Explanations}

For an item $x$ the contribution of feature $j$ is its Shapley value with respect to background items $b$
of the same selection (in the system, 100 items sampled among those with $\pi<\tau/2$),
\begin{equation}
\phi_j(x)=\sum_{S\subseteq F\setminus\{j\}}\frac{|S|!\,(d-|S|-1)!}{d!}\;\Delta_j(S),
\end{equation}
\begin{equation}
\Delta_j(S)=v\bigl(S\cup\{j\}\bigr)-v(S),\qquad v(S)=\mathbb{E}_b\, f\bigl(x_S,\,b_{\bar S}\bigr),
\end{equation}
where $F$ is the set of $d=19$ features, $\bar S=F\setminus S$, and $f(x_S,b_{\bar S})$ is the
calibrator's prediction for a point that takes the features in $S$ from $x$ and the remaining ones from a
background item $b$. The values are additive: they sum to the difference between the item's
risk and the mean risk of the background. Exact computation requires $2^{19}$ coalitions per item; they are
estimated by Monte Carlo over random feature permutations~\cite{strumbelj2014explaining} (24
permutations per item in the system). Contributions are reported in risk points and labelled in
the auditor's language (e.g.\ ``strength of deviation from the group median'', ``description differs
from the rest of the group'').

\section{System implementation}\label{sec:system}

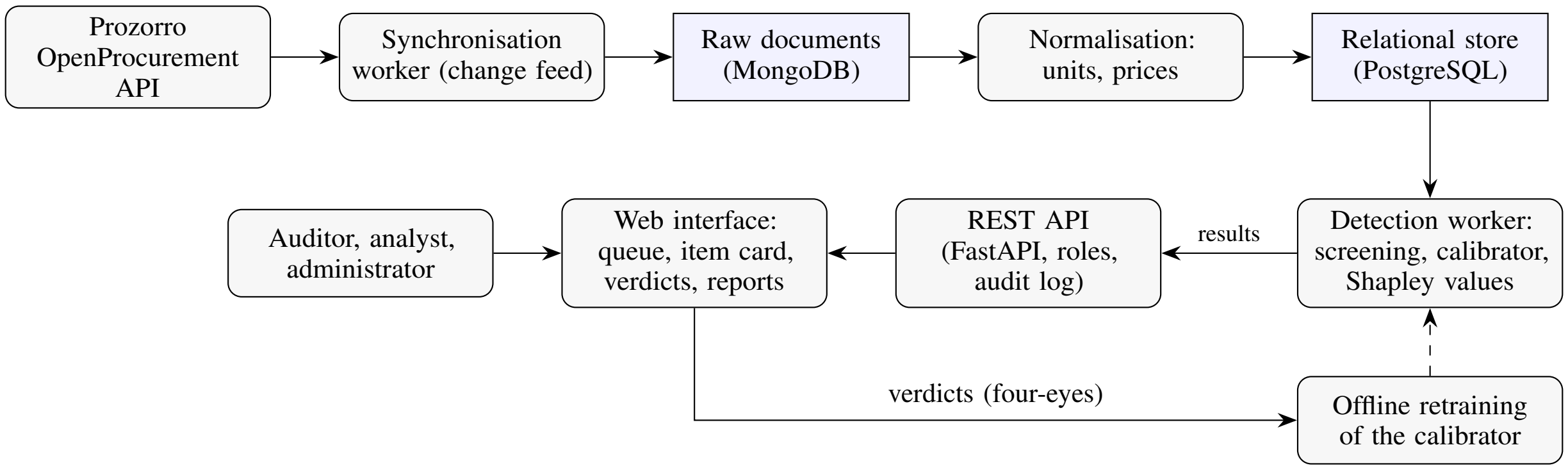
\begin{figure*}[t]
\centering
\begin{tikzpicture}[node distance=5mm and 7mm, every node/.style={font=\footnotesize},
  box/.style={draw, rounded corners, align=center, minimum height=9mm, text width=25mm, fill=gray!6},
  store/.style={draw, align=center, minimum height=9mm, text width=22mm, fill=blue!5},
  arr/.style={-{Stealth[length=2mm]}}]
\node[box] (api) {Prozorro\\OpenProcurement API};
\node[box, right=of api] (sync) {Synchronisation\\worker (change feed)};
\node[store, right=of sync] (mongo) {Raw documents\\(MongoDB)};
\node[box, right=of mongo] (norm) {Normalisation:\\units, prices};
\node[store, right=of norm] (pg) {Relational store\\(PostgreSQL)};
\node[box, below=10mm of pg] (det) {Detection worker:\\screening, calibrator,\\Shapley values};
\node[box, left=14mm of det] (rest) {REST API\\(FastAPI, roles,\\audit log)};
\node[box, left=of rest] (ui) {Web interface:\\queue, item card,\\verdicts, reports};
\node[box, left=of ui] (users) {Auditor, analyst,\\administrator};
\node[box, below=7mm of det] (train) {Offline retraining\\of the calibrator};
\draw[arr] (api) -- (sync); \draw[arr] (sync) -- (mongo); \draw[arr] (mongo) -- (norm); \draw[arr] (norm) -- (pg);
\draw[arr] (pg) -- (det);
\draw[arr] (det) -- node[above, font=\scriptsize]{results} (rest);
\draw[arr] (rest) -- (ui); \draw[arr] (users) -- (ui);
\draw[arr] (ui.south) |- node[pos=0.75, above]{verdicts (four-eyes)} (train.west);
\draw[arr, dashed] (train) -- (det);
\end{tikzpicture}
\caption{Architecture of the information system. Solid arrows: data flow; dashed: the retrained
calibrator replaces the default model after its training set has been reviewed.}
\label{fig:architecture}
\end{figure*}

The method runs inside an information system for audit staff (Fig.~\ref{fig:architecture}). A
synchronisation worker follows the Prozorro change feed and stores raw tender documents in a document
store (MongoDB); a normalisation step maps them into a relational schema (PostgreSQL) of tenders, lots,
items, contracts, procuring entities and suppliers, derives unit prices and converts units. Database
triggers keep the base-unit price consistent with the item record. An interrupted synchronisation resumes
from the last processed change; one day of the feed (about 21 thousand changed documents) took about nine
minutes.

An auditor defines a selection by CPV codes, period and units; before the run the system reports a
homogeneity index of the selection (the spread of its categories), because detection is more precise in
homogeneous selections. Detection runs in a separate worker: for the reference selection of 16{,}103 items
the full pipeline takes about 38~s, or about 45~s including Shapley values for the candidates. Results are cached by
selection and parameters, so a repeated request returns immediately.

Each analysed item receives a risk score, the decision of the calibrator, the data-error indicator, the
group it was compared with and, for candidates, the feature contributions. The review queue lists the
union of high-risk items and likely data errors and can be filtered by either signal. A reviewer opens
an item card with the plain-language explanation, the contributions and comparable purchases, and
records a verdict (confirmed, not anomalous, needs review) with a reason; a verdict becomes final only
after approval by a second auditor (four-eyes principle). Verdicts, approvals and reports are kept with
their history, and changes of users and parameters are recorded in an audit log by database triggers.
Three roles separate duties: analysts synchronise data and may override detection parameters,
auditors review items, and administrators manage users and default parameters.

Under a load test with 100 concurrent simulated users following auditor and analyst scenarios, run with
all services and the load generator on one Docker host with 4 virtual CPUs and 3.8~GB of memory, the
system served about 48 requests per second without errors with a 95th-percentile latency of
1.8--1.9~s; indexing the queue, the item card and the item lists reduced the corresponding query times by
factors of 16--36. The code base is covered by 154 automated tests and is published under the AGPL-3.0
licence.

\section{Evaluation design}\label{sec:eval}

\paragraph{Stratified annotation} True anomalies are rare, so a simple random sample would contain few
positives. We stratified the reference selection by the decision of the first system version (v1): the 50
highest-risk flagged items (\emph{top}), the remaining 577 flagged items (\emph{mid}) and 14{,}823 unflagged
analysed items (\emph{normal}); 50, 100 and 250 items were annotated respectively. Population metrics are
estimated with Horvitz--Thompson weights $w_h=N_h/n_h$~\cite{horvitz1952sampling}; confidence intervals use a
stratified bootstrap with 2{,}000 replicates~\cite{efron1993bootstrap}.

\paragraph{Annotation protocol} An item is \emph{confirmed} if an auditor should examine its unit price:
a substantial deviation from comparable goods not explained by the description (as a guide, above
$\sim$2$\times$ or below $\sim$0.4$\times$ the typical price), a price that clearly refers to another unit
(pack, gram, whole lot), or another evident data error. Items whose description does not allow a judgement
are marked \emph{cannot judge} and excluded; excluding them assumes that they are not informative about
the other labels. Both annotation rounds were performed by a large language model (Claude, Anthropic)
under this written protocol, which served as the instruction. The sighted round was produced in
September 2026 in two waves (125 and 275 cards, batches of 25--55; model version not recorded). The blind
round was produced on 24~November~2026 by Claude Opus~5.5 in eight batches of 50 cards. It was not an
isolated annotator: the same session had access to the project repository, knew the purpose of the study,
had seen aggregate statistics of the sighted round (but not its per-item labels while annotating) and had
built the blind cards itself; only the cards' content was restricted to the item record and neutral
references. The human check of 80 items (Section~\ref{sec:res-labels}) was performed by the first author,
who is not a professional auditor, on the same blind cards with the model's labels hidden. The protocol (in
Ukrainian and English), all cards and labels are published with the code; prompts beyond the protocol were
not used and no transcripts are published. Table~\ref{tab:examples} shows typical cases of each label.

\begin{table*}[t]
\caption{Examples from the blind annotation (descriptions translated from Ukrainian). Reference: median
price of items with the same eight-digit CPV code and unit ($n$ items).}
\label{tab:examples}
\centering
\footnotesize
\begin{tabular}{@{}lrrrll@{}}
\toprule
Item description & Price in record & Quantity & Reference (n) & Label & Reason \\
\midrule
Butter ``Selianske'' & 470 UAH per \emph{gram} & 0.8 & 400 UAH/kg (694) & confirmed & unit error: price per kg, unit gram \\
Butter & 83.6 UAH/kg & 320 & 400 UAH/kg (694) & confirmed & unit error: price of a 200 g pack per kg \\
Pangasius fillet, frozen & 231{,}800 UAH/kg & 1 & 190 UAH/kg (914) & confirmed & data error: lot total as unit price \\
Sunflower oil, refined & 0.11 UAH/kg & 4{,}600 & 93 UAH/kg (354) & confirmed & data error: implausible unit price \\
Sugar & 140 UAH/kg & 50 & 35 UAH/kg (619) & confirmed & over-priced, $4\times$ a homogeneous good \\
Tomato paste 25\%, 480 g & 141.75 UAH/jar & 16 & 59 UAH/jar (30) & confirmed & over-priced, $2.4\times$ \\
Pork & 35 UAH/kg & 286 & 210 UAH/kg (662) & confirmed & under-priced, $0.17\times$ \\
Whole-grain pasta & 115 UAH/kg & 120 & 45 UAH/kg (609) & not anomalous & dearer variety explains the price \\
Oil (4600 g) & 385 UAH/piece & 12 & 95 UAH/piece (154) & not anomalous & 5-litre bottle; references are 1-litre \\
Various food products & 42.31 UAH/unit & 153 & 431 UAH/unit (34) & cannot judge & content of the ``unit'' unknown \\
\bottomrule
\end{tabular}
\end{table*}

\paragraph{Sighted and blind rounds} In the first (\emph{sighted}) round, annotation cards displayed the
v1 decision and risk score together with the median and three comparable items of the pipeline's own
group; the cards also carried the stratum and the price ratio and were ordered by stratum, the top stratum
by decreasing risk. Because this can anchor the annotator on the model's output, we re-annotated all 400 items
\emph{blind}: cards show only the item record and neutral reference prices (median, 10th--90th percentile
and eight random examples of items with the same CPV code and unit, drawn directly from the database rather
than from the pipeline's groups; for three items the code-level references were too few and the five- or
four-digit category was used), in shuffled order. The rounds therefore differ in more than the display
of the decision: the reference information, the card format and the annotation session differ as well. All
results below use the blind labels unless stated otherwise.

\paragraph{Evaluation of the ranking model} The calibrator is evaluated with nested cross-validation: five outer
folds repeated five times; in each outer training set the decision threshold is chosen by an inner
five-fold cross-validation maximising design-weighted F1. Reported calibrator scores therefore come only
from models that have not seen the item and whose threshold was not tuned on it. The bootstrap intervals
resample items with these predictions fixed; they do not include the variance of refitting the model, so
they understate the total uncertainty. Thirteen annotated items that fall into groups too small for v2 are
scored 0.

\paragraph{Baselines and alert budget} We compare with 3$\sigma$ and Tukey rules on log prices within
CPV class $\times$ unit (an auditor without the system) and within the eight-digit code, and with
Isolation Forest, LOF and ECOD applied (i) to log prices within the pipeline's groups and (ii) to the 19
calibrator features without labels. Since unsupervised detectors have no natural threshold, we compare them
at an equal alert budget of 4\% of analysed items (the share flagged by v1) and by average precision (AP),
which does not depend on a threshold. Detectors run within groups return the percentile rank of an
item in its group, which makes scores comparable across groups but gives the top item of every group the
same score; this handicaps their ranking across groups and hence their AP.

\paragraph{Synthetic errors} Annotations measure what an annotator considers relevant; they cannot
measure recall for error types that are rare in a 400-item sample. We therefore inject errors with known
ground truth into 3\% of items, in equal shares of four types typical of procurement records:
\emph{moderate} price deviations ($\times2$, $\times3$, $\times0.4$, $\times0.5$), \emph{strong} deviations
($\times5$, $\times10$, $\times0.1$, $\times0.2$), \emph{unit-scale} errors ($\times1000$ or $\times0.001$,
grams vs.\ kilograms, kilograms vs.\ tonnes) and \emph{lot totals} entered as unit prices (price multiplied
by the quantity, for items with quantity $\ge5$). We repeat the injection with three seeds in CPV~15 and in
the held-out CPV~03 and apply the calibrators without retraining. Genuine anomalies already present in the
data are unlabelled and count as false positives, so precision is a lower bound and methods that flag fewer
items are favoured in precision and F1; recall by error type is exact. The Isolation Forest baseline is
given the alert budget of the calibrator trained on sighted labels (6.5\% in CPV~15).

\section{Results}\label{sec:results}

\subsection{Annotation agreement and anchoring}\label{sec:res-labels}

\begin{figure}[t]
\centering
\includegraphics[width=\columnwidth]{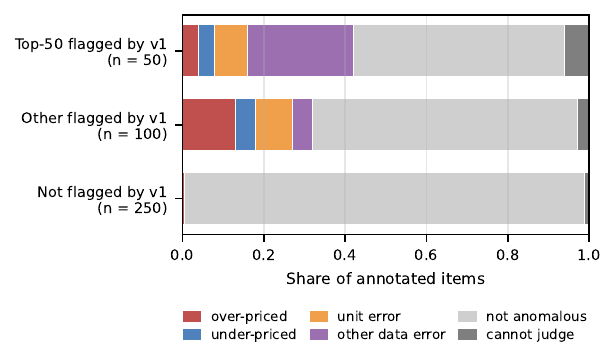}
\caption{Blind labels by stratum of the stratified sample. Data errors (wrong unit, lot totals,
implausible values) are concentrated among the highest-risk items of v1; over- and under-pricing is more
common among the other flagged items; a single confirmed case occurs among the 247 judged unflagged
items.}
\label{fig:labels}
\end{figure}

\begin{figure*}[t]
\centering
\includegraphics[width=0.78\textwidth]{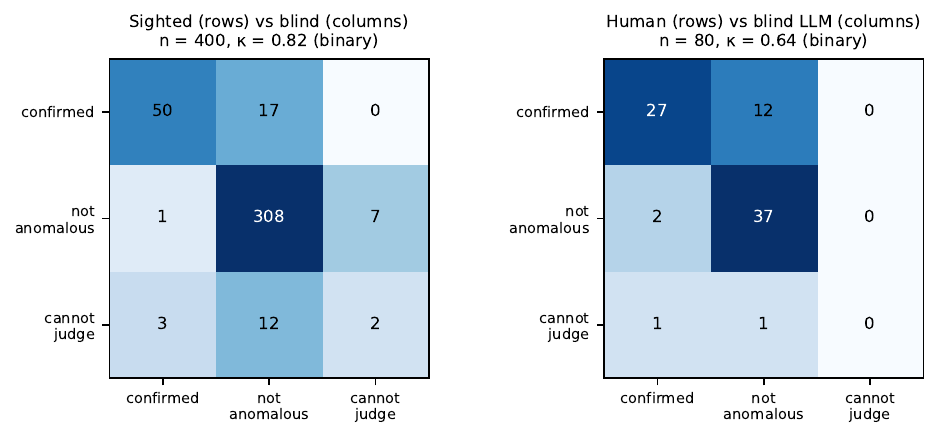}
\caption{Agreement between annotation rounds (all items shown; $\kappa$ computed on the items judged in
both, $n=376$ and $n=78$). Left: the sighted round (annotator saw the v1 decision) against the blind round;
the 17 items confirmed only by the sighted round are mostly items flagged by v1.
Right: the human check against the blind language-model annotation; the human confirmed 12 items the model
did not, mostly goods whose reference prices were not comparable.}
\label{fig:agreement}
\end{figure*}

Of the 400 items, the blind annotation confirmed 54 and marked 9 as \emph{cannot judge}
(Fig.~\ref{fig:labels}). Over half of the confirmed items are data errors rather than price deviations:
unit errors (a price per kilogram entered for a gram, or the price of a 200~g pack entered per kilogram)
and other data errors such as lot totals entered as unit prices. For an audit body both kinds matter ---
the first can hide or simulate over-pricing, the second distorts published spending statistics.
Agreement between
the sighted and the blind annotation is high (Cohen's $\kappa=0.82$~\cite{cohen1960kappa} on the 376 items
judged in both rounds; $\kappa=0.68$ including the \emph{cannot judge} class), but the disagreements are not
symmetric. Among items flagged by v1, the sighted round confirmed 14 items that the blind round did not,
against 4 in the opposite direction (3 of which the sighted round had marked \emph{cannot judge});
confirmation rates fell from 52\% to 42\% in the top stratum and from 37\% to 32\% in the mid stratum.
Among unflagged items the sighted round also confirmed more items (4 vs.\ 1), which anchoring on an
unflagged decision would not predict and which points to a generally more lenient sighted round; it also
used \emph{cannot judge} twice as often (17 vs.\ 9). The shift among flagged items is
consistent with anchoring on the displayed decision, but because the rounds also differ in reference
information and session, it cannot be attributed to anchoring alone (Fig.~\ref{fig:agreement}, left).

A human annotator (not a professional auditor) labelled a subsample of 80 blind cards (30 confirmed and 50
not confirmed by the blind annotation, order shuffled, annotation hidden). On the 78 items both judged,
raw agreement was 82\% and $\kappa=0.64$. Disagreements were again asymmetric: the human flagged 12 items
the model did not, against 2 in the opposite direction. Most of the 12 are goods whose neutral reference
prices are not truly comparable --- vanilla sugar shown next to ordinary sugar, dry yeast next to fresh
yeast, whole-grain pasta, juice in 200~ml packs --- where the human followed the reference prices literally
and the model adjusted for the product. Comparability thus limits not only automatic screening but also
human judgement, which supports the emphasis of Section~\ref{sec:method} on the comparability layer.

\subsection{Detection quality}

\begin{table*}[t]
\centering
\caption{Design-weighted quality on blind labels (reference selection, CPV 15). P, R, F1 at the method's
own threshold or at the 4\% alert budget; AP is threshold-free.}
\label{tab:labels}
\small
\begin{tabular}{lcccc}
\toprule
Method & P & R & F1 & AP \\
\midrule
3$\sigma$, CPV class $\times$ unit & 0.489 & 0.427 & 0.456 & 0.506 \\
Tukey (IQR), CPV class $\times$ unit & 0.274 & 0.645 & 0.384 & 0.455 \\
Tukey (IQR), CPV code $\times$ unit, 4\% budget & 0.331 & 0.623 & 0.432 & 0.449 \\
Isolation Forest in pipeline groups, 4\% & 0.413 & 0.711 & 0.522 & 0.395 \\
LOF in pipeline groups, 4\% & 0.219 & 0.715 & 0.336 & 0.176 \\
ECOD in pipeline groups, 4\% & 0.245 & 0.776 & 0.372 & 0.285 \\
Isolation Forest on calibrator features, 4\% & 0.150 & 0.405 & 0.219 & 0.153 \\
ECOD on calibrator features, 4\% & 0.171 & 0.492 & 0.254 & 0.176 \\
\midrule
v1 (screening only) & 0.339 & 0.780 & 0.473 & 0.493 \\
\quad + pack-size normalisation & 0.473 & 0.952 & 0.632 & 0.544 \\
\quad + finer semantic split (v2 screening) & 0.478 & 0.952 & 0.637 & 0.544 \\
v2 screening without CPV grouping & 0.405 & 0.423 & 0.414 & 0.382 \\
v2 screening without semantic split & 0.320 & 0.602 & 0.418 & 0.358 \\
\midrule
\textbf{v2: screening + monotone calibrator (nested CV)} & \textbf{0.637} & \textbf{0.796} & \textbf{0.708} & \textbf{0.808} \\
Hybrid: extremes always flagged + calibrator & 0.500 & 0.865 & 0.634 & 0.612 \\
\bottomrule
\end{tabular}
\end{table*}

Table~\ref{tab:labels} summarises the results; AP denotes the design-weighted average precision. Three
findings stand out. First, some form of grouping is essential, but a simple rule already captures much of
it: the 3$\sigma$ rule on CPV class $\times$ unit reaches AP 0.51, the full screening stage 0.54; removing
CPV grouping or the semantic split from the screening stage lowers its AP to 0.38 and 0.36, and pack-size
normalisation raises AP from 0.49 to 0.54. (The F1 values of the screening variants at their own
thresholds, 0.47 to 0.64, mostly reflect those thresholds: at the common 4\% budget their F1 is 0.59 in
all three cases.) Second, by AP generic unsupervised detectors do not beat simple robust rules on this
problem: within the same groups, Isolation Forest, LOF and ECOD reach AP of 0.18--0.40; applied
to the 19 calibrator features without labels they perform worst (AP 0.15--0.18), because those features are
informative only in combination with the annotator's notion of relevance. By F1 at the 4\% budget,
however, Isolation Forest within groups (0.52) does better than the 3$\sigma$ rule (0.46) and v1 at its own
threshold (0.47). Third, the learned ranking is the
largest single improvement: AP rises to 0.81, a gain of 0.31 over v1 (95\% interval $[0.17, 0.43]$), 0.28
over the 3$\sigma$ rule ($[0.10, 0.44]$) and 0.24 over the screening stage ($[0.06, 0.40]$). The gain in F1
at the selected threshold (0.47 $\to$ 0.71) is less certain, $[-0.01, 0.44]$, because the threshold is chosen
on few positives in inner folds. The \emph{hybrid} variant, which flags every extreme regardless of the
score, lowers AP to 0.61: the annotation does not consider every extreme price relevant, often because the
good is not comparable with its group. This gap rests largely on two unflagged items (weight 60 each)
that are extremes but were labelled not anomalous; without them, AP is 0.82 for the ranking model and 0.74
for the hybrid. On the sighted labels the conclusions are unchanged (AP 0.47 $\to$
0.75, F1 0.47 $\to$ 0.75).

The recall component of these estimates rests on few items in the large unflagged stratum: its single
confirmed item carries 22\% of the weighted positives. Without it, AP is 0.79 for the ranking model, 0.55
for the 3$\sigma$ rule and the screening stage and 0.49 for v1, so the ordering does not depend on this
item.

\begin{figure}[t]
\centering
\includegraphics[width=0.85\columnwidth]{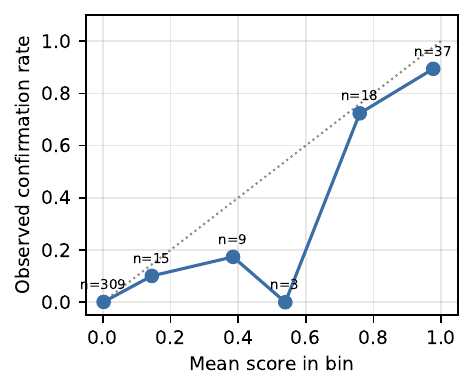}
\caption{Reliability of the ranking scores (nested cross-validation, design-weighted). Points: mean score
and observed confirmation rate per score bin; numbers: items per bin.}
\label{fig:reliability}
\end{figure}

The scores are reasonably reliable at both ends of the scale --- items scored above 0.9 are confirmed in
89\% of the weighted cases, items below 0.1 in 0.1\% --- but over-estimate the confirmation rate in the
middle (Fig.~\ref{fig:reliability}); the design-weighted Brier score is 0.007, against 0.017 for a constant
prediction of the base rate. We therefore use the scores for ranking and thresholding, not as probabilities.

\begin{figure}[t]
\centering
\includegraphics[width=\columnwidth]{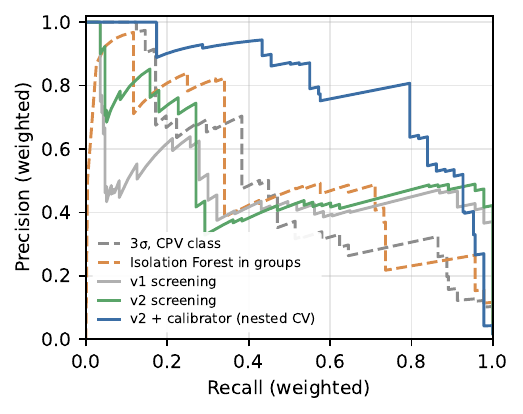}
\caption{Design-weighted precision--recall curves on blind labels. The ranking model keeps precision
at about 0.75 or higher up to a recall of 0.6, where the screening stage and the baselines are below 0.5.}
\label{fig:pr}
\end{figure}

Fig.~\ref{fig:pr} shows the same comparison over all thresholds. The ranking model dominates the
screening stage and the baselines in the high-precision region that matters for an audit queue: an auditor
works through the top of the list, and every false alarm there costs review time. At high recall the
curves converge, because the remaining confirmed items are those that no method ranks highly.

\begin{figure*}[t]
\centering
\includegraphics[width=0.85\textwidth]{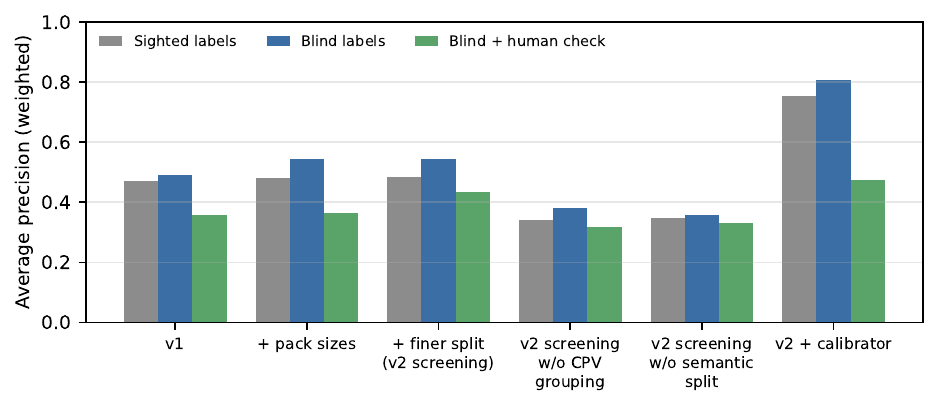}
\caption{Average precision of the screening stage, its ablations and the ranking model under the three
label variants. In all three, removing CPV grouping or the semantic split lowers AP of the screening stage
by 0.11--0.19. The gain from the ranking model is large on the language-model labels and small on the
labels with human corrections.}
\label{fig:ablation}
\end{figure*}

\paragraph{Sensitivity to the annotator} Substituting the human labels on the 80 verified items changes
the picture considerably: 63 instead of 54 items are confirmed, most of the additions are goods whose
reference prices are not comparable, and the labels now mix two standards. Six of the twelve human-only
confirmations fall into the unflagged stratum, of which the human checked 36 of 247 items; the seven
confirmed unflagged items carry 65\% of the weighted positives. The screening-based methods lose quality
(v1: AP 0.36; screening stage of v2: AP 0.44), whereas some detectors gain (ECOD within groups: AP 0.29
$\to$ 0.30, F1 0.37 $\to$ 0.44; Isolation Forest on the calibrator features: AP 0.15 $\to$ 0.35). That the
human confirmed six of 36 unflagged items the language model rejected suggests that, under human-like
labels, the population recall of all methods would be much lower than the blind labels indicate. The ranking model remains the best ranker by AP (0.48), but its F1 (0.43) is no better than that
of ECOD within the pipeline's groups (0.44), and its AP gains are no longer significant: 0.12 over v1
(95\% interval $[-0.04, 0.33]$) and 0.09 over the 3$\sigma$ rule ($[-0.09, 0.30]$). The benefit of the learned ranking is therefore tied to a consistent notion of relevance
in the labels it is trained and evaluated on, while the effect of the grouping ablations is similar across
label variants (Fig.~\ref{fig:ablation}).

\subsection{Synthetic errors and a held-out category}

\begin{table*}[t]
\centering
\caption{Injected errors (3\% of items, three seeds; mean). Precision is a lower bound; R by type is
recall for moderate, strong, unit-scale and lot-total errors. Calibrators are trained on CPV~15 labels and
applied without retraining; CPV~03 is held out. The best F1 in each category is in bold (ties included).}
\label{tab:synthetic}
\small
\setlength{\tabcolsep}{4pt}
\begin{tabular}{llccccccc}
\toprule
 & Method & F1 & AP & Flagged & R mod. & R strong & R unit & R lot \\
\midrule
\multirow{9}{*}{CPV 15}
 & 3$\sigma$, CPV class & \textbf{0.58} & 0.63 & 1.7\% & 0.01 & 0.18 & 0.92 & 0.70 \\
 & Isolation Forest in groups & 0.51 & 0.42 & 6.1\% & 0.50 & 0.80 & 0.90 & 0.87 \\
 & v1 & 0.52 & 0.62 & 6.1\% & 0.48 & 0.80 & 0.95 & 0.92 \\
 & v2 screening & 0.55 & 0.62 & 5.4\% & 0.49 & 0.79 & 0.90 & 0.91 \\
 & v2, calibrator on sighted labels & 0.45 & 0.46 & 6.5\% & 0.49 & 0.74 & 0.80 & 0.85 \\
 & v2, monotone calibrator on blind labels & 0.57 & 0.49 & 3.6\% & 0.33 & 0.68 & 0.74 & 0.80 \\
 & hybrid, same calibrator & \textbf{0.58} & 0.55 & 4.4\% & 0.38 & 0.74 & 0.89 & 0.89 \\
 & v2, monotone calibrator on blind + human labels & 0.40 & 0.43 & 8.7\% & 0.54 & 0.81 & 0.86 & 0.90 \\
 & hybrid, same calibrator & 0.40 & 0.53 & 9.1\% & 0.56 & 0.84 & 0.91 & 0.93 \\
\midrule
\multirow{9}{*}{CPV 03}
 & 3$\sigma$, CPV class & 0.61 & 0.65 & 1.5\% & 0.01 & 0.05 & 0.97 & 0.83 \\
 & Isolation Forest in groups & 0.56 & 0.50 & 5.4\% & 0.53 & 0.77 & 0.93 & 0.91 \\
 & v1 & 0.61 & 0.67 & 4.7\% & 0.43 & 0.80 & 0.99 & 0.95 \\
 & v2 screening & 0.61 & 0.68 & 4.4\% & 0.43 & 0.77 & 0.94 & 0.92 \\
 & v2, calibrator on sighted labels & 0.50 & 0.59 & 5.7\% & 0.47 & 0.76 & 0.83 & 0.89 \\
 & v2, monotone calibrator on blind labels & \textbf{0.66} & 0.63 & 3.1\% & 0.34 & 0.67 & 0.83 & 0.85 \\
 & hybrid, same calibrator & \textbf{0.66} & 0.65 & 3.7\% & 0.39 & 0.74 & 0.93 & 0.92 \\
 & v2, monotone calibrator on blind + human labels & 0.47 & 0.57 & 6.5\% & 0.49 & 0.76 & 0.87 & 0.90 \\
 & hybrid, same calibrator & 0.48 & 0.63 & 7.0\% & 0.53 & 0.81 & 0.94 & 0.94 \\
\bottomrule
\end{tabular}
\end{table*}

\begin{figure*}[t]
\centering
\includegraphics[width=0.85\textwidth]{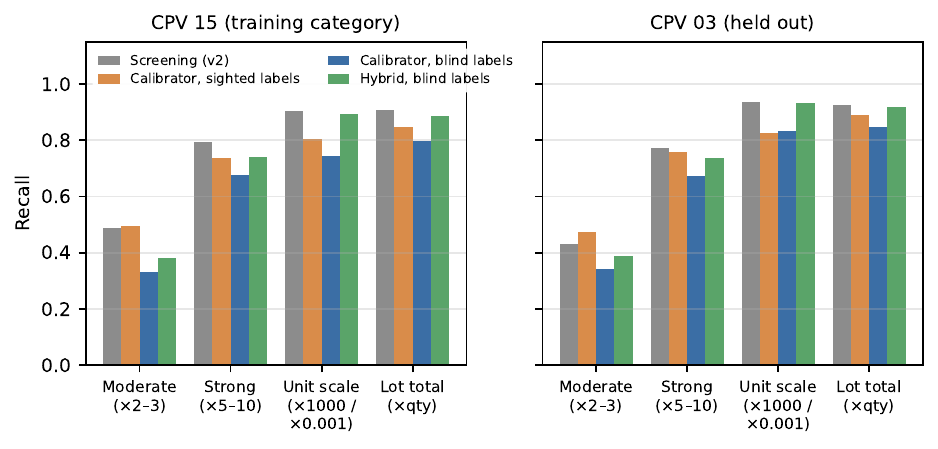}
\caption{Recall by injected error type (mean of three seeds). The blind-label calibrator misses part of
the unit-scale errors; forcing extremes to be flagged (hybrid) restores it.}
\label{fig:synthetic}
\end{figure*}

Table~\ref{tab:synthetic} and Fig.~\ref{fig:synthetic} report the injected-error benchmark. Unit-scale errors and lot totals are
found by all pipeline variants and by simple rules; moderate deviations ($\times2$, $\times3$, $\times0.4$, $\times0.5$) are found by none
reliably. The calibrator trained on the sighted labels lowers F1 below the screening stage in both
categories. The monotone calibrator trained on blind labels is the most conservative (3--4\% of items
flagged); in the held-out CPV~03 it has the highest F1 (0.66 vs.\ 0.61 for screening), but this advantage
comes from flagging fewer items, which the benchmark rewards, and its AP (0.63) is below that of the
screening stage (0.68) and of the 3$\sigma$ rule (0.65). It thus remains usable in a category it was not
trained on, but ranks injected errors worse than screening; it also misses a quarter of unit-scale errors
in CPV~15. The hybrid recovers them (recall 0.74 $\to$ 0.89 in CPV~15, 0.83 $\to$ 0.93 in CPV~03) at
unchanged F1. On injected errors the hybrid is therefore preferable, whereas on annotated relevance
(Table~\ref{tab:labels}) the pure calibrator ranks better --- the two evaluations measure different jobs, a
point we return to in the Discussion. The AP of all ranking-model variants is below that of the screening
stage on injected errors, because the calibrator ranks injected moderate deviations lower than the
annotator-relevant cases it was trained on.

\subsection{Explanations}\label{sec:res-explain}

\begin{figure}[t]
\centering
\includegraphics[width=\columnwidth]{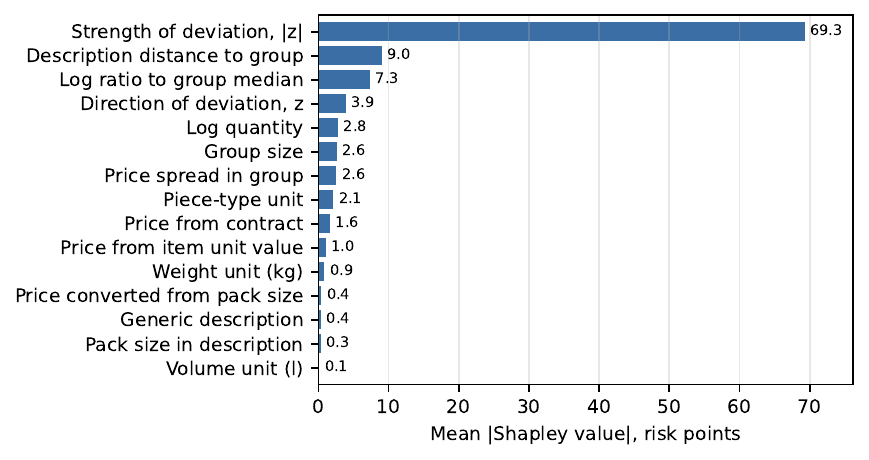}
\caption{Global feature importance of the default calibrator: mean absolute Shapley value over 120
flagged items, in risk points (features with less than 0.1 point omitted).}
\label{fig:importance}
\end{figure}

We explained 120 randomly chosen flagged items of the reference selection (292 flagged by the default
monotone calibrator) against 100 background items (unflagged items with a risk $100\,\pi$ below 30) and
compared the system's estimator (24 permutations) with a reference estimate from 2{,}000 permutations
(Table~\ref{tab:explain}). The error is averaged over all 19 features, several of which have near-zero
attributions, and the additivity gap is reported for the reference estimate.

\begin{table}[t]
\centering
\caption{Quality of Monte Carlo Shapley explanations (120 flagged items, 10 seeds; risk in points of
the 0--100 scale).}
\label{tab:explain}
\footnotesize
\begin{tabular}{@{}p{0.66\columnwidth}r@{}}
\toprule
Mean absolute error vs.\ 2{,}000-permutation reference & 0.98 \\
Top-3 features shared with the reference & 85.0\% \\
Spearman correlation of $|\phi|$ with the reference & 0.95 \\
Top-3 features shared between seeds & 79.5\% \\
Median additivity gap of the reference, $|\sum_j\phi_j-(f(x)-\mathbb{E}_b f(b))|$ & 0.02 \\
Time per item, ms & 32 \\
\midrule
Risk drop, top-1 / 2 / 3 features replaced by background & 87.9 / 89.7 / 91.1 \\
Risk drop, 1 / 2 / 3 random features replaced & 9.7 / 13.0 / 17.1 \\
\bottomrule
\end{tabular}
\end{table}

The error of the 24-permutation estimate is about one risk point on a 0--100 scale and the attribution
magnitudes correlate strongly with the reference ($\rho=0.95$). The three features shown first agree with
the reference in 85\% and across seeds in 80\% of cases; disagreements concern the second and third
features, which often have similar small contributions --- the first-ranked feature is stable. The
explanations are faithful in the deletion sense: replacing the most important feature by background
values lowers the risk by 88 points on average, against 10 points for a random feature; because a single
feature dominates, this test mainly confirms that the explanation identifies it. Globally (Fig.~\ref{fig:importance}, mean
$|\phi|$), the strength of the deviation from the group median dominates (69 points), followed by the
textual distance of the description to its group (9), the log price ratio (7) and the direction of the
deviation (4); quantity, group size, spread and unit type contribute 2--3 points each. The textual
distance raises the risk on average (+8 points; it lowers it for only 8\% of flagged items): an item whose
description differs from the rest of its group is more often confirmed, typically because it is a
different good priced as if it were comparable, or a mis-coded item.

\section{Discussion}

\paragraph{Two signals, not one} The experiments separate two jobs that a single anomaly score
conflates. Gross data errors --- unit-scale confusions and lot totals entered as unit prices --- are found
by the screening stage alone, without labels, in both categories (recall 0.90--0.99). Deciding which of
the many deviations deserve an auditor's time is where the learned ranking adds value (AP 0.49 $\to$
0.81 on annotated relevance), and that value is defined by the annotation it learns from. Merging the two
into one decision trades one against the other: the hybrid restores recall of unit-scale errors but
lowers AP on annotated relevance from 0.81 to 0.61 (0.82 vs.\ 0.74 without the two items discussed in
Section~\ref{sec:results}). The system therefore presents two signals rather than a single
score: a learned \emph{risk} and a \emph{likely data error} indicator, raised for every extreme of the
screening stage regardless of the calibrator. The review queue contains the union of both, can be filtered
by either, and the item card explains a likely data error with what to check (unit of measure, pack size,
lot total entered as a unit price). In a selection of three CPV classes (1553 butter, 1551 milk, 1542 refined oils; 2{,}318 analysed
items), 49 items have high risk and 42 are likely data errors; 15 of the latter would not have reached the
queue through the calibrator alone.

\paragraph{Labels are part of the model} Three annotation variants of the same 400 items --- sighted,
blind, and blind with human corrections --- lead to different calibrators. The model trained on sighted
labels flags more items and ranks injected errors no better than the blind-label model; the human corrections,
made by a non-expert who followed reference prices literally, made the calibrator flag more than twice as
many items (8.7\% vs.\ 3.6\% of CPV~15 items in the synthetic benchmark), partly because cross-validation
selected a lower threshold for it (0.30 vs.\ 0.675). For an audit organisation this means that the annotation protocol, the information shown to
annotators and the composition of the reviewer pool are design decisions of the detector, and that
verdicts collected in production should be audited before they are used for retraining.

\paragraph{Grouping first, then learning} In all three label variants, removing CPV grouping or the
semantic split lowers AP of the screening stage by 0.11--0.19, and generic unsupervised detectors do not
outperform robust rules within well-formed groups. Grouping is necessary but simple: a class-level
3$\sigma$ rule comes within 0.04 AP of the full screening stage. The large gain comes from learning which
deviations matter, which in turn depends on comparable groups to compute its features.

\paragraph{Operating point} On the reference selection the ranking model flags 1.9\% of the analysed items
(292 of 15{,}034), the v2 screening stage 3.6\% and v1 4.1\%; a queue of about 300 items for 15 thousand
analysed items; the data-error indicator adds the extremes that the calibrator ranks low (15 of 64 queued
items in the selection discussed above). Because the ranking quality of the
learned model is high at the top of the list (Fig.~\ref{fig:pr}), an audit unit with limited capacity
can work through the queue from the top and stop when its budget is spent; the threshold matters less than
the ordering. The threshold can be adjusted by analysts within bounds; we report F1 at the threshold chosen
by cross-validation only to compare methods.

\subsection{Threats to validity}

\paragraph{Construct validity} ``Anomalous price'' is operationalised as what a reviewer following the
written protocol considers worth an auditor's attention. This is neither a legal finding nor evidence of
a violation, and the protocol's guide values ($2\times$, $0.4\times$) are a convention. The human check
shows that a different reviewer applies the protocol differently when reference prices are not comparable.
The synthetic benchmark measures detection of known error types, not their relevance.

\paragraph{Internal validity} Both annotation rounds were produced by the same language-model family; the
agreement between them is therefore an upper bound on the agreement of independent annotators. The blind
round removed the model's decision from the cards, but the neutral reference prices are themselves a form
of guidance. The same model family also assisted in designing and implementing the detector and the
experiment code, so the labels and the method may share blind spots. The strata were defined by the v1
decision; Horvitz--Thompson weighting keeps the estimates unbiased, but the recall estimate rests on few
confirmed items in the large \emph{normal} stratum (one in the blind round, weight 60). The ranking model's
threshold is chosen on few positives, and the bootstrap does not include refitting variance. Researcher
degrees of freedom remain: the screening thresholds were tuned on injected anomalies in CPV~15, and design
choices of the ranking model (the monotone constraints, the per-pack feature) were informed by the same 400
labels that nested cross-validation later evaluates; nested cross-validation does not protect against
such choices.

\paragraph{External validity} The data cover one procurement system, a change-feed window of two weeks
(most items published on 10--24 September 2026) and two CPV divisions of goods with physical units; seasonal
effects are not covered. Services, works and goods priced per set or per service unit are
less comparable and were not evaluated. Transfer of the calibrator was tested only on injected errors in
the second category, not on its annotated relevance.

\paragraph{Responsible use} A list of suspicious prices names procuring entities. Several safeguards
follow from the design: the system shows its reasons (the comparison group, the deviation and the feature
contributions) rather than a bare score; a verdict requires a second auditor's approval; the data-error
indicator is worded as a request to check the record, not as an accusation; and the interface does not
use the term ``violation''. Results should be used to prioritise review, not to sanction.

\section{Conclusion}
We proposed and evaluated a method for screening unit prices of individual items in public procurement,
in which a comparability layer and robust screening find deviations, a monotone ranking model trained on
annotated verdicts orders them by how likely an annotator is to confirm them, and gross extremes are
reported separately as likely data errors, with per-item Shapley explanations. On Prozorro food purchases,
the learned ranking clearly improves design-weighted average precision over statistical rules, generic
unsupervised detectors and the screening stage on language-model labels; explanations computed with 24
permutations are accurate and stable. Our evaluation also shows that the labels are the weakest link: a
sighted annotation round confirmed the earlier system's flags more often than a blind round, and with the
corrections of a human annotator the gain of the learned ranking shrinks and is no longer significant. Future work should replace language-model annotation with verdicts of
professional auditors collected blind, extend the evaluation to more categories and periods, and study how
auditors use the explanations in their decisions.

\section*{Data and code availability}
Prozorro data are public and available through the OpenProcurement API~\cite{prozorroapi}. The
annotation cards, all three annotation variants, the experiment code and result tables are available at
\url{https://github.com/DmytroBabarytskyi/tendoor} (directory \texttt{research}; code under AGPL-3.0,
annotation data under CC BY 4.0).

\section*{Declaration on Generative AI}
During the preparation of this work, the authors used Claude (Anthropic) in order to: annotate the
evaluation sample as described in Section~\ref{sec:eval};
write the experiment scripts of Section~\ref{sec:results}, the code that produces
Figs.~\ref{fig:labels}--\ref{fig:importance} from the recorded results;
paraphrase and format the manuscript. The research design, the detection method, the information system
and the conclusions are the authors' own. Every reported number was checked against the
published result files and the system's database. After using this tool, the authors reviewed and edited
the content as needed and take full responsibility for the content of the publication.


\end{document}